\documentclass{article}
\usepackage{spconf,amsmath,graphicx,leftidx}
\usepackage{amsfonts}
\usepackage{amssymb}
\usepackage{amsthm,amsmath}
\usepackage{mathrsfs}
\usepackage{indentfirst}

\usepackage{graphicx}
\usepackage{subfigure}
\usepackage{multirow}
\usepackage{xcolor}
\usepackage{booktabs}

\title{SGE Net: VIDEO OBJECT DETECTION WITH SQUEEZED GRU AND INFORMATION ENTROPY Map
}
\name{Rui Su$^{\star}$ \qquad Wenjing Huang$^{\star}$ \qquad Haoyu Ma$^{\dagger}$ \qquad Xiaowei Song$^{\ddagger}$ \qquad Jinglu Hu$^{\star}$}
  
\address{$^{\star}$ Waseda University, Fukuoka, Japan\\
$^{\dagger}$University of California, Irvine, USA\\
$^{\ddagger}$Southeast University, Nanjing, China}

\begin{document}
%
\maketitle
\begin{abstract}
Recently, deep learning based video object detection has attracted more and more attention. 
Compared with object detection of static images, video object detection is more challenging due to the motion of objects, while providing rich temporal information. The RNN-based algorithm is an effective way to enhance detection performance in videos with temporal information. However, most studies in this area only focus on accuracy while ignoring the calculation cost and the number of parameters. 

In this paper, we propose an efficient method that combines channel-reduced convolutional GRU (\underline{S}queezed \underline{G}RU), and Information \underline{E}ntropy map for video object detection (\textbf{SGE-Net}). The experimental results validate the accuracy improvement, computational savings of the Squeezed GRU, and superiority of the information entropy attention mechanism on the classification performance. The mAP has increased by 3.7 contrasted with the baseline, and the number of parameters has decreased from 6.33 million to 0.67 million compared with the standard GRU.

\end{abstract}
%

\section{Introduction}
\label{sec:intro}
Object detection plays a fundamental role in computer vision, as it usually performs as the first step of several downstream tasks such as segmentation\cite{he2017mask, tang2019nodulenet} and pose estimation\cite{sun2019deep, kong2020sia, chen2020nonparametric, wang2019geometric}. 
Compared with detection on static images~\cite{Ren2015FasterRT}, detection in videos has much more application scenarios, such as robot navigation and intelligent video surveillance.
The naive way of object detection in videos is applying an image-based detector on each frame independently. However, compared with static images, the properties of objects in video, such as appearance, shape, etc., will change with the motion of the objects. This will introduce more challenges like motion blur. 
Thus, an image-based detector usually does not generalize well on video data and it may fail on some intermediate frames in videos~\cite{Xiao2018VideoOD}. 
Meanwhile, running the detector frame by frame is time-consuming and requires a lot of computing resources, which limits its application in real scenes.

Compared with static images, videos contain temporal information. The time cue provides both low-level and high-level correspondences between frames~\cite{wang2019learning}, and these correspondences can help inference based on previous frames. 
Traditionally, researchers applied tracks~\cite{Han2016SeqNMSFV} or optical flow ~\cite{Zhu2017DeepFF, Hetang2017ImpressionNF} to model the correspondences in video. However, these methods can only exploit short-term correspondence.  
Recently, there has been lots of works exploit recurrent network, such as RNN, GRU and LSTM, in video object detection. 
Specifically, there are two groups of methods, one is postprocessing method~\cite{Donahue2015LongtermRC, Lu2017OnlineVO, Ning2017SpatiallySR}, the other is integrating method~\cite{Xiao2018VideoOD, Liu2018MobileVO}. 
However, RNN-based modules require lots of parameters, which slows down the inference time. 
To address these limitations, in this paper, we introduce the channel-reduced convolutional GRU, named \textit{Squeezed GRU}. Compared with the standard GRU and convolutional GRU~\cite{siam2017convolutional}, our model can efficiently extract the spatial-temporal correlations in video with just 10\% amount of parameters of the standard GRU, while without loss of detection performance.

Meanwhile, Convolutional Neural Networks (CNN) can only capture local features due to the limited receptive fields. Thus it may ignore the overall perception of the entire image. 
On the other hand, traditional image processing algorithms such as histogram and entropy of images can provide a global evaluation, but cannot provide any semantic or spatial cues. 
To solve these issues, we propose a novel 2D \textit{Information Entropy} map to capture additional semantic cues without additional learnable parameters, and demonstrate that it can further enhance the feature representation learned by CNN.

To summarize,  the main contributions of our work are: 
\begin{enumerate}
\vspace{-0.5em}
    \item We propose the channel-reduced convolutional GRU, named squeezed GRU, to efficiently model the temporal information across frames with few parameters. 
    \vspace{-0.5em}
    \item We propose the 2D Information Entropy map to capture semantic cues of an entire image without supervision.
    \vspace{-0.5em}
    \item We proposed the SGE-Net, which combines both the squeezed GRU and Information Entropy map, and demonstrate its effectiveness on video object detection.

\end{enumerate}

\begin{figure}[h]
\begin{minipage}[b]{1.0\linewidth}
  \centering
  \includegraphics[width=8.5cm]{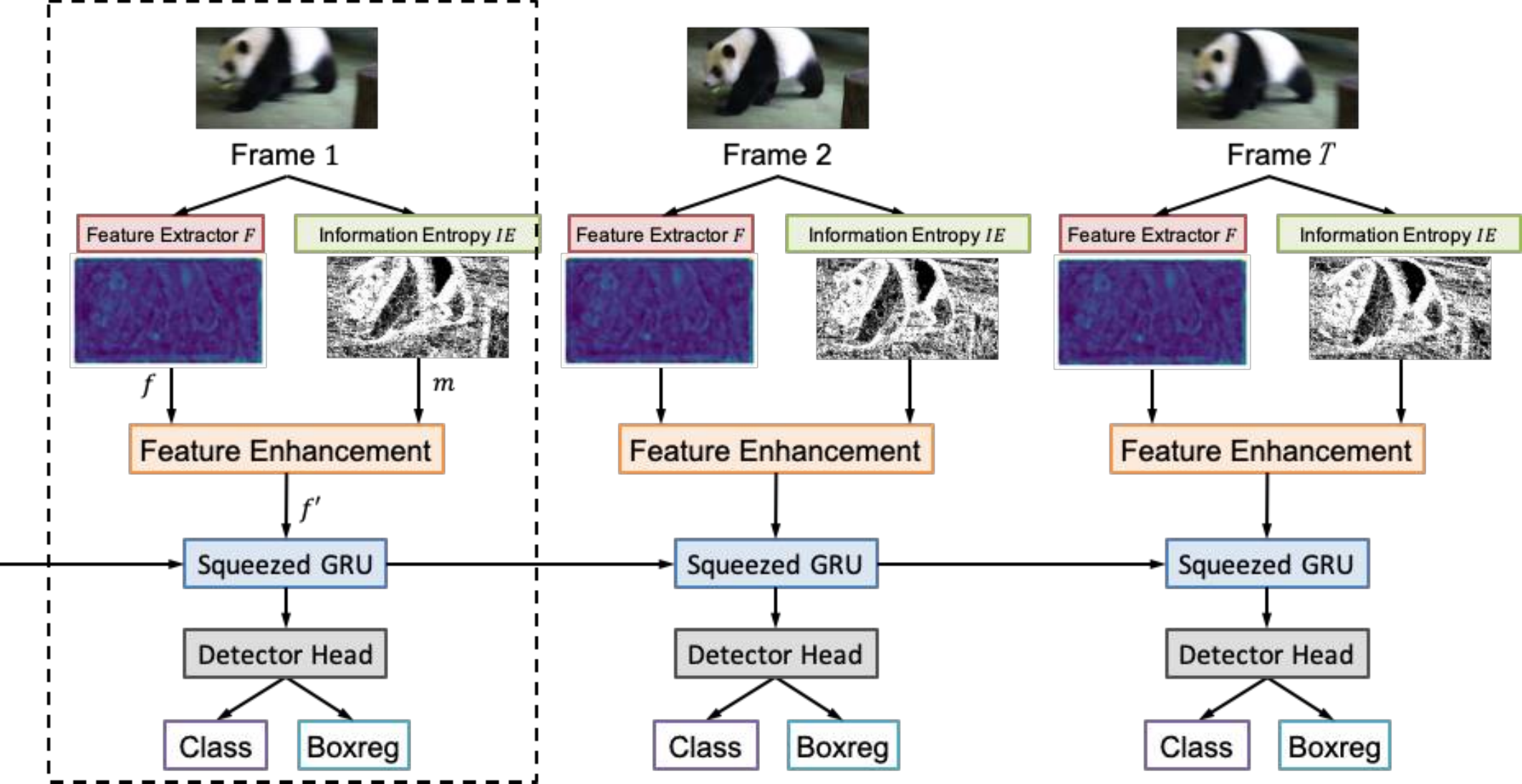}
\end{minipage}
\vspace{-0.5em}
\small 
\caption{The proposed network architecture. }
\label{fig:model}
\end{figure}

\section{Methodology}
\label{sec:format}
\vspace{-0,5em}
\subsection{Architecture}
\vspace{-0,5em}
The overall framework of our proposed \textit{SGE-Net} is shown in Figure \ref{fig:model}. For each frame $I \in \mathcal{R}^{H \times W \times 3}$ of a video clip, it firstly goes through a feature extractor $\mathcal{F}(\cdot)$ to obtain the intermediate feature map $f=\mathcal{F}(I) \in \mathcal{R}^{h \times w \times c} $. Here $c$ is the number of channels, and $h$, $w$ are the height and weight of the feature map, respectively. Meanwhile, the Information Entropy module $\mathcal{IE}(\cdot)$ extracts the entropy map $m \in \mathcal{R}^{h \times w}$ from the original image $m = \mathcal{IE}(I)$. Then the feature enhancement module refines the feature $f$ with $m$, and output the refined feature $f'=\mathcal{FE}(f, m)$. Furthermore, to fully exploit the temporal correlation, we introduce the Squeezed GRU module on top of the refined feature $f'$. Finally, the detector head outputs the bounding box and the category.


\vspace{-0.5em}
\subsection{Information Entropy (IE) Map}
\vspace{-0.5em}

\textbf{Rationale. } 
Given a grayscale image $I$, denote the histogram of $I$ as $H_I = \{p_i | {i = 0,1,2,..,C}\}$, where $p_i$ is the proportion of pixels whose value is $i$, and $C$ is the maximum pixel value and is usually equal to $255$ for 8-bit grayscale image. The histogram of an image can reflect the tonal distribution of pixel values while cannot imply spatial and semantic information about the image.  
\textit{Information Entropy} (IE) can reflect the uncertainty of information, including images.
In general, the more information presented in the image, the larger entropy value it will have. Based on this, we propose two \textit{hypotheses} about IE for the detection tasks: 
(1) Different objects may hold significantly different IE values. The object and environment usually hold different IE values. 
(2) Different ranges of IE correspond to different object categories. Thus the distribution of IE for different categories may vary. 
Thus, the IE of an image may imply some semantic cues, which may help strengthen the feature map learned by the neural network. \\
\textbf{Implementation. }
The standard unary entropy value of an image $I$ can be calculated by: 
\begin{equation}
h_{I} = -\sum_{i=0}^{C}p_{i}\log{p_{i}}
\end{equation}
where $p_i$ and $C$ is defined from the histogram $H_I$. 

However, this standard entropy is one-dimensional, and it can only represent the amount of information of the entire image $I$, while cannot reflect the spatial characteristics. 
Furthermore, different parts of one image may contain different objects and thus hold varied entropy.
The single number entropy $h_I$ cannot represent this information as well. 
To address these issues and characterize the spatial information, we propose a novel  \textit{two-dimensional Information Entropy}: 

\begin{gather}
p_{ij} = \frac{f(i,j)}{N^2}\\
h_{2D} = -\sum_{i=0}^{C}\sum_{j=0}^{C}p_{ij}\log_2{p_{ij}}
\end{gather}

where $i$ is pixel gray value and $j$ is the average grayscale of all pixels in the sliding window. $f(i,j)$ is the frequency of this binary group. $N$ is the number of pixels in a sliding window. 
 $p_{ij}$ denotes the probability of the binary group, which reflects the comprehensive characteristics of the gray value at a pixel location and the gray distribution of the surrounding pixels. Binary entropy $h_{2D}$ is calculated by accumulating $i$ and $j$ from 0 to $C=255$ in 8-bit grayscale image.
Then we divide the image $I$ into multiple windows and apply the sliding window to the entire picture to calculate their binary entropy. 
We set the window size to 3 and operate twice to generate an entropy map $m = \mathcal{IE}(I)$. 
According to the above formulas, the first sliding can generate the same number of duals as the number of pixels, and the second sliding can obtain the same number of entropy as the number of pixels. 
Through two operations of the sliding window, each pixel in the information entropy map can 
model the relationship between pixels and gray distribution, and contain the long term correlation.





\begin{figure}[htb]
\small
\begin{minipage}[b]{.25\linewidth}
  \centering
  \centerline{\includegraphics[width=2.5cm]{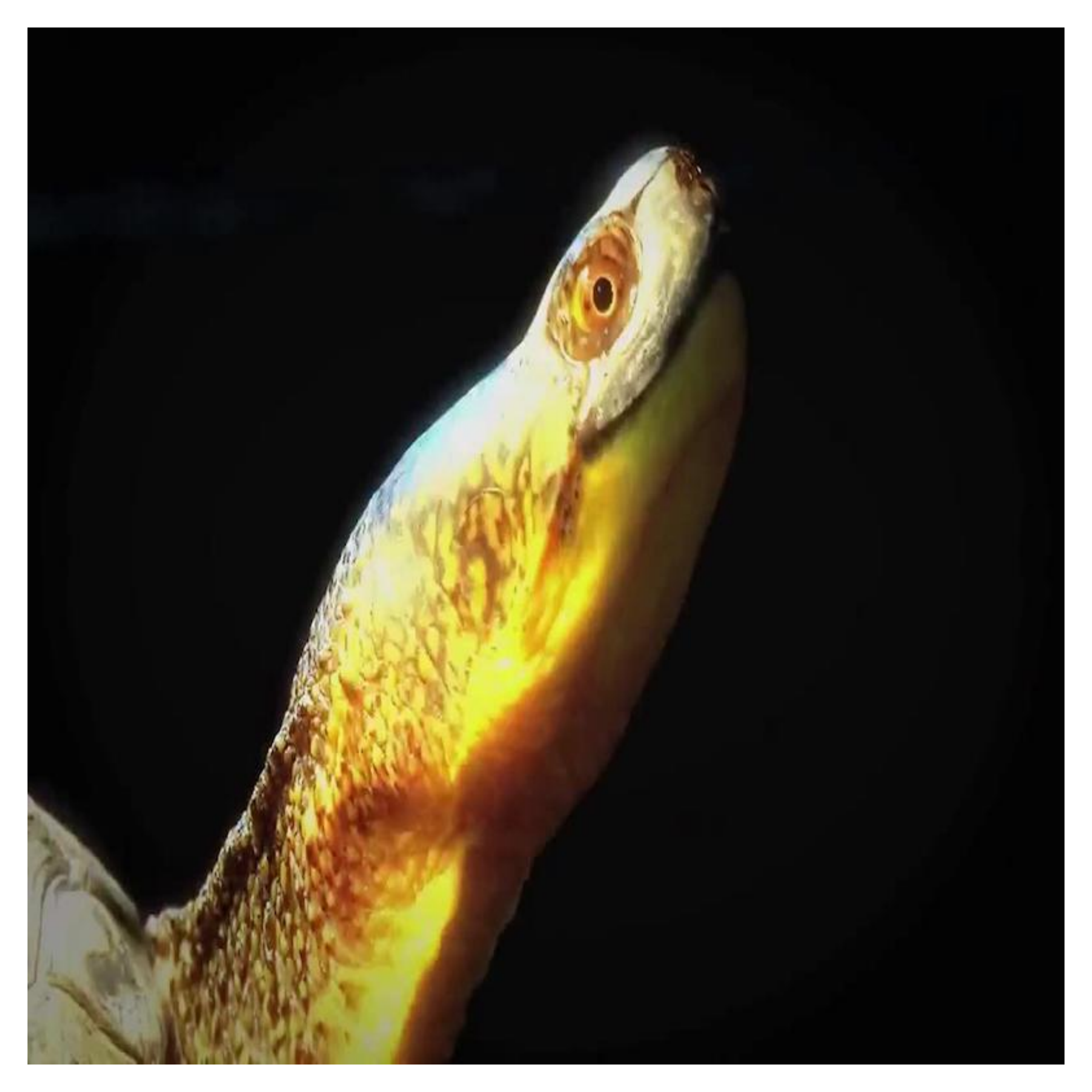}}
\end{minipage}
\hfill
\begin{minipage}[b]{.25\linewidth}
  \centering
  \centerline{\includegraphics[width=2.5cm]{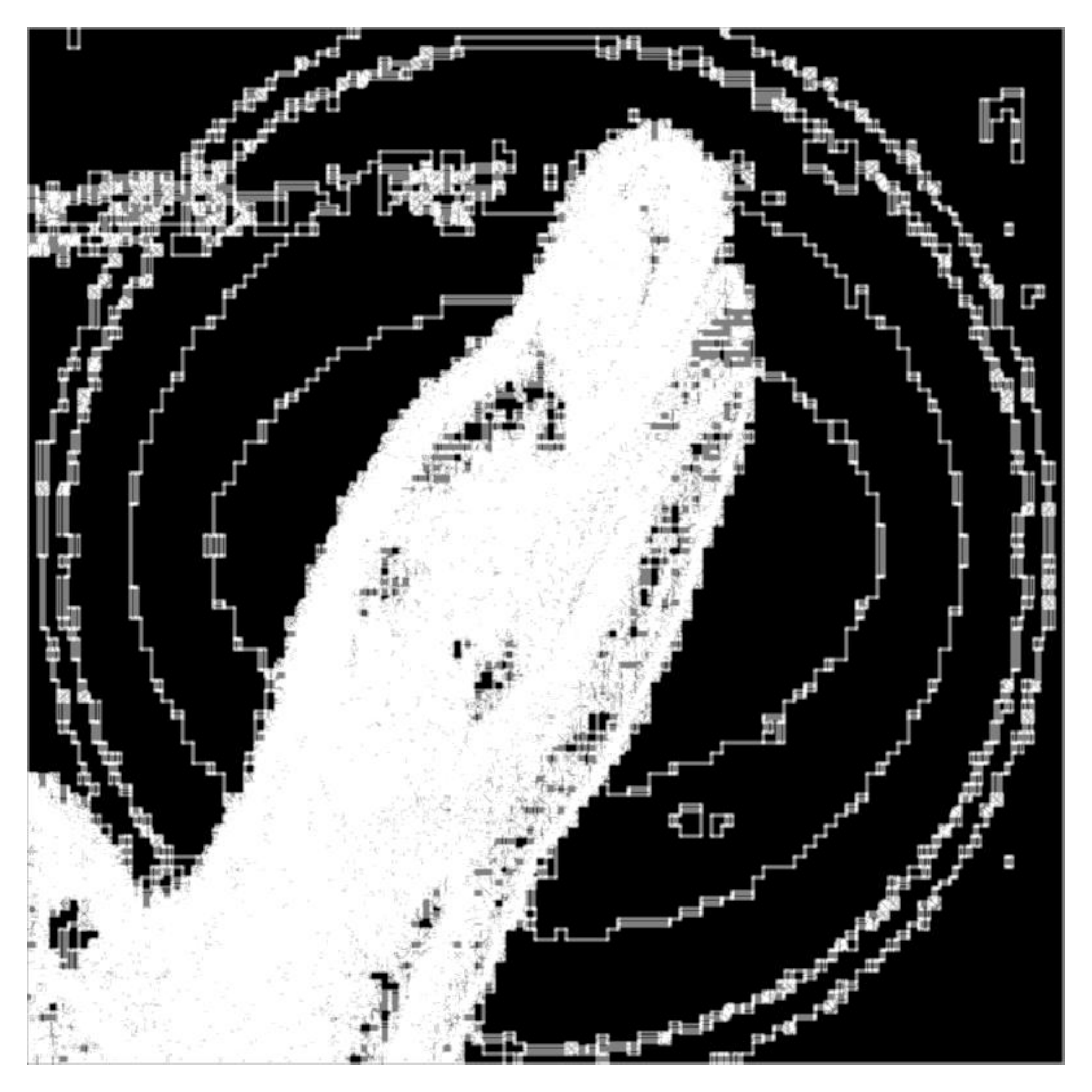}}
\end{minipage}
\hfill
\begin{minipage}[b]{.35\linewidth}
  \centering
  \centerline{\includegraphics[width=4cm]{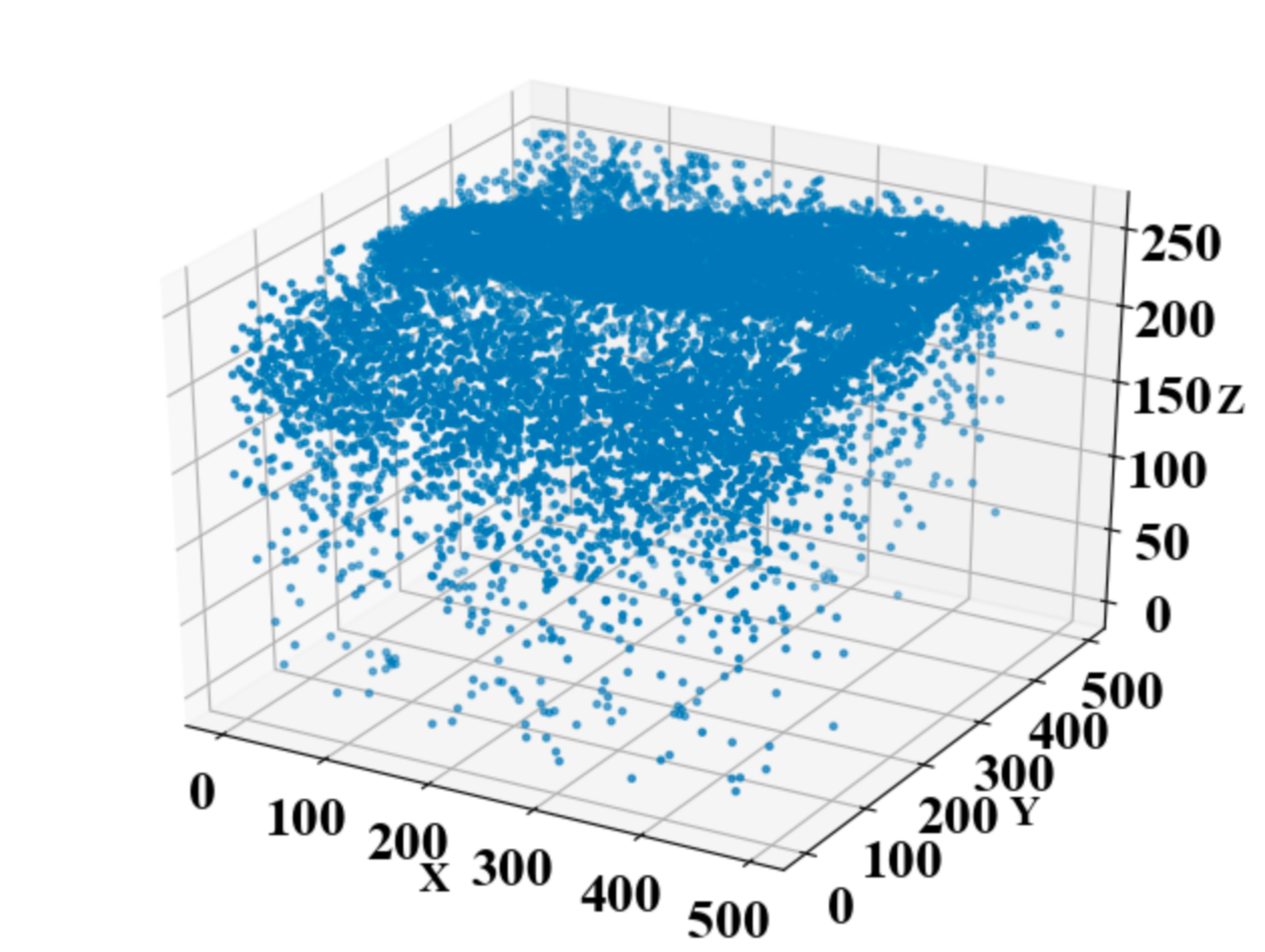}}
\end{minipage}

\vspace{-0.5em}
\begin{minipage}[b]{0.25\linewidth}
  \centering
  \centerline{\includegraphics[width=2.5cm]{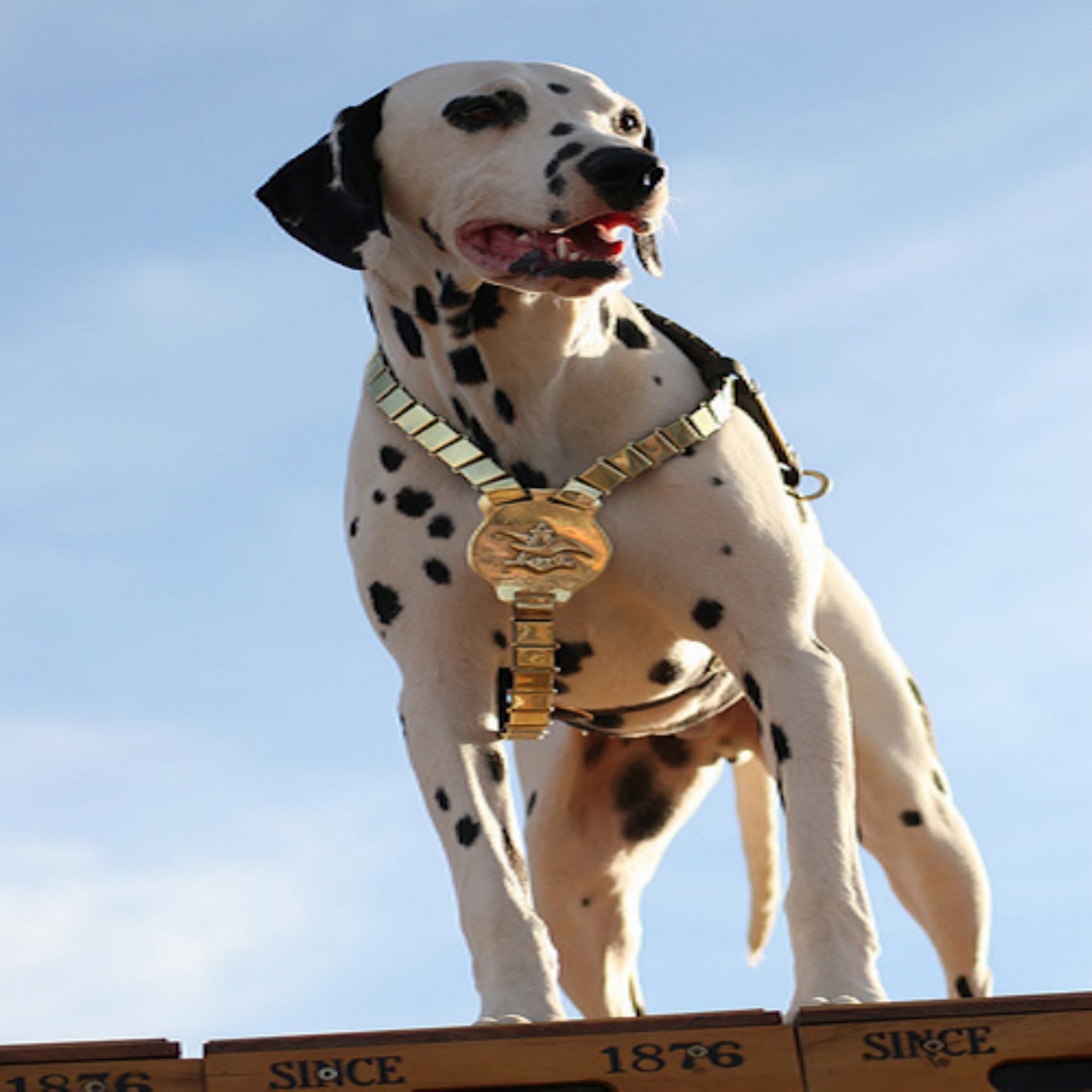}}
\end{minipage}
\hfill
\begin{minipage}[b]{.25\linewidth}
  \centering
  \centerline{\includegraphics[width=2.5cm]{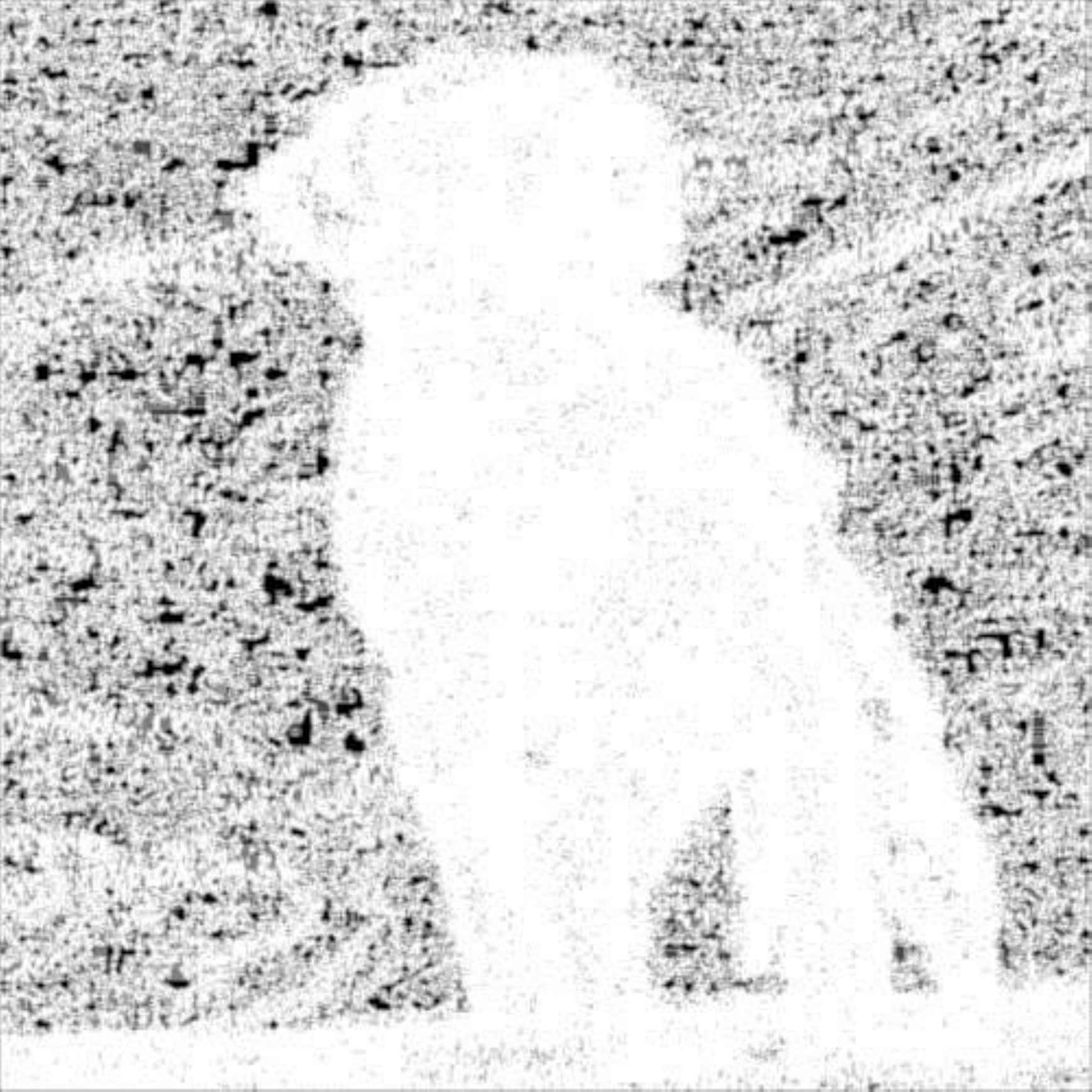}}
\end{minipage}
\hfill
\begin{minipage}[b]{.35\linewidth}
  \centering
  \centerline{\includegraphics[width=4cm]{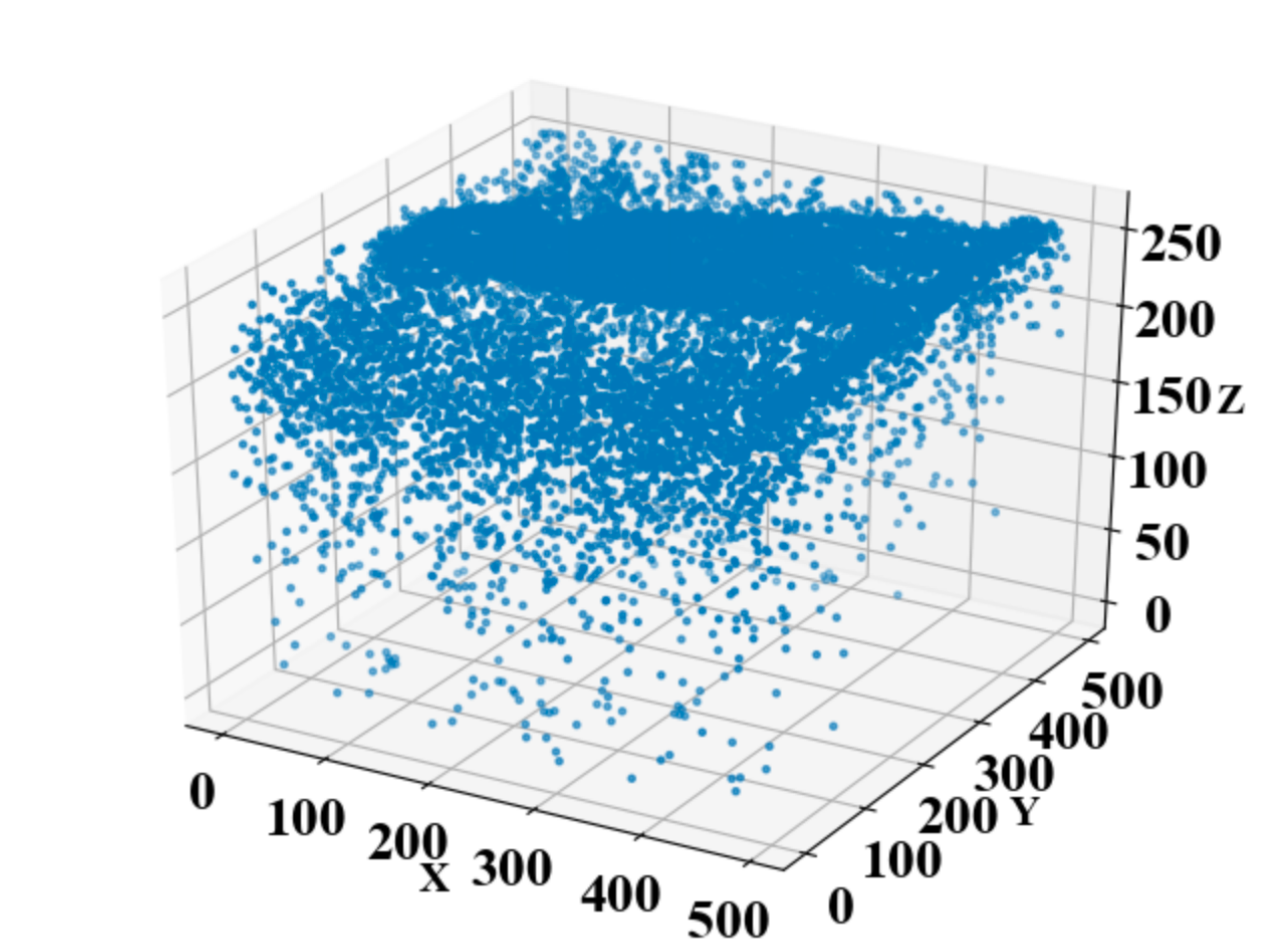}}
\end{minipage}

\vspace{-0.5em}
\caption{ \small{Examples of information entropy maps. The three columns are the raw images, the entropy map displayed in 2D, and the entropy map displayed in 3D, respectively.}}
\label{fig:turtle}
\end{figure}
Figure \ref{fig:turtle} shows examples of our proposed information entropy map based on the original image. To enhance the independence of objects and prevent the adhesion of pixels, we use the open operation in morphology to help generate the map. From the 3D display, we notice that the object and the environment have different 2D entropy values, thus our IE map may help locate the object in the image. Furthermore, different objects usually hold different entropy value distributions, thus the IE map may also help classify the objects.   \\
\textbf{Feature Enhancement.}
Once we obtain the IE map $m \in \mathcal{R}^{H\times W} $, we can apply it to refine any intermediate feature map $f$ with the Feature Enhancement module: $\mathcal{FE(\cdot, \cdot)}$
\begin{equation}
    f' = \mathcal{FE}(f, m) =  \sigma( \text{pooling}(m)) * f  \in \mathcal{R}^{h \times w \times c}
\end{equation}
where $\sigma$ is the sigmoid function aiming to scale the IE map to 0 and 1, and \textit{pooling} is the pooling layer aiming to reduce the size of IE map. Thus, the IE map $m$ can serve as an attention module to locate the objects and their categories.

\vspace{-0.5em}
\subsection{Squeezed GRU}
\vspace{-0.5em}
GRU \cite{chung2014empirical} and LSTM \cite{hochreiter1997long} are two variants of Recurrent Neural network. Compared with LSTM, GRU has a simpler structure and fewer matrix multiplications, thus it is relatively efficient.
Nevertheless, the standard GRU with Hadamard products is not applicable in video detection. The image is regarded as a vector, which makes them lose the spatial correlation. Also, the Hadamard products require a lot of calculation. 

\begin{figure}[htb]
 \centering
\begin{minipage}[b]{0.8\linewidth}
\includegraphics[width=7cm]{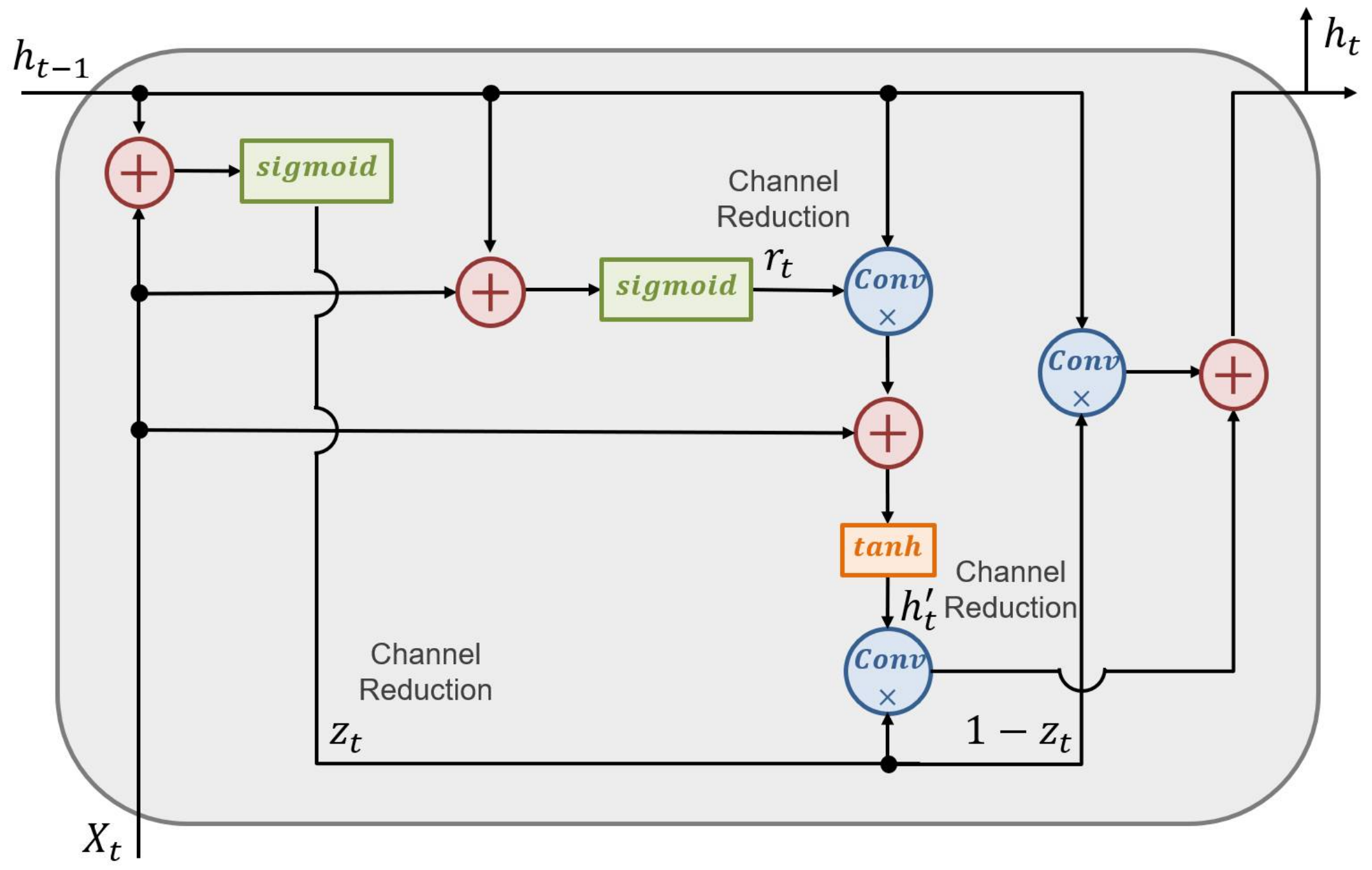}
\end{minipage}
\vspace{-0.5em}
\small
\caption{The structure of squeezed GRU. The Hadamard products of normal GRU are replaced by depthwise separable convolutions, where they are represented by the symbol $conv$. ($1\times1$) convolutions are used to reduce the number of channels, where they are marked with $Channel$ $Reduction$. Here the number of channels is reduced from ($1024+256$) to $256$.}
\label{fig:GRU}
\end{figure}

To solve these limitations of GRU, we proposed the channel-reduced Convolutional GRU, named \textit{Squeezed GRU}. It has two improvements compared with the standard GRU. .
(1) We replace the Hadamard products with convolutional operation, as the convGRU \cite{siam2017convolutional}. In this way, the network has the ability to simultaneously learn the temporal information in the video and keep the spatial information in the hidden state. 
(2) We use the \textit{depth-wise separable convolution} proposed by Howard et al.~\cite{Howard2017MobileNetsEC} instead of the standard convolutional operation in convGRU. Specifically, we reduce the channel three times and name it as squeezed operations. 
With these two characteristics, we can greatly reduce the calculation amount when calculating hidden states and input vectors, and maintain the spatial information efficiently. 
The architecture of the \textit{squeezed GRU} is illustrated in Figure \ref{fig:GRU} and is formulated by: 
\begin{align}
{z_t} &= \theta{(\leftidx{^{In+Out}{W}{_z^{Out}}} \times [{X_t},h_{t-1}] )}\\
{r_t} &= \theta{(\leftidx{^{In+Out}{W}{_r^{Out}}} \times [{X_t},h_{t-1}] )}\\
{h_{t}^{'}} &= \varphi{(\leftidx{^{In+Out}{W}{_h^{Out}}} \times [{X_t},{r_t}\circ h_{t-1}] )}\\
{h_t} &= (1-{z_t}) \circ {h_{t-1}}+{z_t}\circ {h_{t}^{'}}
\end{align}

where $ X_t $ is the feature map input to the squeezed GRU, $ h_ {t-1} $ is the hidden state tensor, and $ h_t $ is the output tensor of this layer. $\theta$ and $\varphi$ mean sigmoid activation function and tanh activation function respectively. $\circ$ denotes Hadamard production. The superscript ($In+Out$) represents the the input channel and the superscript $Out$ represents the channel after reducing. $[]$ is the symbol that applies concatenation before convolution.

\section{Experiments}
\vspace{-0.5em}
\subsection{Experimental Settings}
\vspace{-0.5em}
\textbf{Networks.}  
We adopt the famous one-stage detector SSD ~\cite{Liu2016SSDSS} as our baseline detector module. 
Moreover, as our method mainly focus on the efficiency, we replace its backbone from the ResNet~\cite{He2016DeepRL} to the ligheweight model, the MobileNet~\cite{Howard2017MobileNetsEC}, and we name it as MNet-SSD for short. We further integrate our proposed information entropy attention and Squeezed GRU into the intermediate layer of the network. For comparison, we also try the MNet-SSD with the standard LSTM, GRU, and convGRU. \\
\textbf{Dataset.} Similar to previous video object detection methods, we conduct our experiments on the ImageNet VID 2015 \cite{Deng2009ImageNetAL}. This dataset contains 3862 snippets for training and 555 snippets for testing. In total, there are 30 categories. \\
\textbf{Training.} We use Pytorch to implement all experiments. Standard data augmentation is applied to all datasets during the training time. The network is initialized with weights pretrained on ImageNet, and optimized with the RMSprop optimizer with initial learning rate 0.0003 and momentum 0.9. The length $T$ of sequence is set to 10. And the loss function follows the setting of SSD which contains both the L1 loss for bounding box regression and the cross entropy loss for classification. 

\vspace{-0.5em}
\subsection{Results}
\vspace{-0.5em}
We present the mAP of detection on the VID validation dataset and the number of parameters for the temporal module (named ``T-params.") of each model in Table \ref{table:All}. Noticeably, comparing with the Standard GRU and convGRU, our Squeezed GRU only requires 0.67 million parameters to model the temporal correlation, while it can still achieve nearly the same mAP. 
By adding the IE map, it can further gain additional 1.1\% mAP (from 45.7\% to 46.8\%), and our IE map will not introduce additional parameters. 
Besides, we present the multiply-adds columns (MACs) of each model to evaluate the efficiency. 
Compared with the baseline without temporal model, our method only introduces 20.6\% additional MACs (from 11.59 to 13.98), but it highly improves the performance. Compared with Conv GRU, it can save more than 40\% computational costs (from 24.16 to 13.98). 
All these experimental results imply that our method is efficient while maintaining the accuracy.

\begin{table}[t]
\small 
\caption{\small{Results on VID Validation Set. The T-params means the number of parameters for the temporal module, and M means million. The MACs means multiply-adds columns, and G means giga. }}
\vspace{-0.5em}
\label{table:All}
\begin{center}
\begin{tabular}{l |c | c |c}
\toprule
Models & mAP / \% &  T-Params /M  & MACs /G\\
\hline
MNet-SSD \cite{Howard2017MobileNetsEC}  & 43.1 & - &\bf{11.59} \\ 
MNet-SSD+LSTM \cite{Liu2018MobileVO} & 46.3 & 8.41 &41.06\\
MNet-SSD+GRU \cite{Liu2018MobileVO} & 46.5 & 6.33 & 35.51\\
MNet-SSD+ConvGRU & 46.9 & 29.49  & 24.16\\
\hline 
Ours (Squeezed GRU) & 45.7 & \bf{0.67} & 12.18\\
Ours + IE map  & \bf{46.8} & \bf{0.67} & 13.98\\
\bottomrule
\end{tabular}
\end{center}
\end{table}

\vspace{-0.5em}
\subsection{Ablation Study}
\vspace{-0.5em}
\subsubsection{The Placement of Squeezed GRU}
\vspace{-0.5em}

To explore the best location of the squeezed GRU module, we conduct ablation studies by deploying it in different layers of the backbone, as shown in Table \ref{table:GRUplace}. 
For example, the ``Conv-3" means that we put it on the feature map produced by the Conv-3 layer of the MobileNet. The ``Feature Map-1" means that the location where the first feature map of SSD involved with the classification and location except for Conv-13. 
We can notice that our Squeezed GRU works best after Conv-13 layer, which is coordinate with common sense. Conv-13 is the last layer for feature extraction and the subsequent layers are used to classify and locate objects. Such semantic-rich information is learned in Squeezed GRU to help the detector achieve localization and classification. 


\begin{table}
\small 
\caption{\small{Ablation study of placement for Squeezed GRU. }}
\begin{center}
\label{table:GRUplace}
\begin{tabular}{cc|cc}
\toprule
Location & mAP & Location & mAP\\
\hline
None & 43.1 & - & -\\
Conv-3 & 42.5 & Feature Map-1 & 45.2\\
Conv-4 & 43.4 & Feature Map-2 & 44.6\\
Conv-12 & 44.8 & Feature Map-3 & 45.2\\
Conv-13 & \bf{45.7} & Feature Map-4 & 44.9\\
\bottomrule
\end{tabular}
\end{center}
\end{table}

\vspace{-0.5em}
\subsubsection{The Placement of Information Entropy Map}
\vspace{-0.5em}
We also conduct ablation studies about the location of the IE map and apply it on different intermediate feature maps along the network. Similarly, we explore features from shallow layers (Conv-3) to deep layers (feature-map 4). As shown in Table \ref{table:IEAplace}, we witness that applying the IE map on Conv-13 works the best. Furthermore, when applying to low-level feature maps like Conv-3 and Conv-4, the IE map will even hurt the performance. This suggests that the IE map can mainly strengthen the high-level semantic feature, which is corresponding to our hypothesis that the IE can imply some semantic cues of the image. We believe our IE map is applicable to other tasks with different backbones.

\vspace{-0.5em}
\begin{table}
\small 
\caption{ \small{Ablation study of location for the IE map. }}
\begin{center}
\label{table:IEAplace}
\begin{tabular}{cc|cc}
\toprule
Location & mAP & Location & mAP\\
\hline
None & 45.7 & - & -\\
Conv-3 & 43.2 & Feature Map-1 & 45.5\\
Conv-4 & 44.3 & Feature Map-2 & 45.1\\
Conv-12 & 45.5 & Feature Map-3 & 45.3\\
Conv-13 & \bf{46.8} & Feature Map-4 & 45.7\\
\bottomrule
\end{tabular}
\end{center}
\end{table}
\vspace{-0.5em}

\section{Conclusion}
\vspace{-0.5em}
In this paper, we proposed the squeezed GRU to efficiently utilize the temporal information within video frames, and a novel 2D information entropy map directly computed from the RGB image to further strengthen features learned by neural network. By combining these two methods together, our model can greatly reduce the number of parameters while still achieve 3.7 improvement on video detection benchmark comparing with the baseline. 
In future work, we hope to apply our squeezed GRU and information attention maps to other video-based tasks like segmentation and pose estimation.

\bibliographystyle{IEEEbib}
\bibliography{strings,refs}

\begin{thebibliography}{10}

\bibitem{he2017mask}
Kaiming He, Georgia Gkioxari, Piotr Doll{\'a}r, and Ross Girshick,
\newblock ``Mask r-cnn,''
\newblock in {\em Proceedings of the IEEE international conference on computer
  vision}, 2017, pp. 2961--2969.

\bibitem{tang2019nodulenet}
Hao Tang, Chupeng Zhang, and Xiaohui Xie,
\newblock ``Nodulenet: Decoupled false positive reduction for pulmonary nodule
  detection and segmentation,''
\newblock in {\em International Conference on Medical Image Computing and
  Computer-Assisted Intervention}, 2019, pp. 266--274.

\bibitem{sun2019deep}
Ke~Sun, Bin Xiao, Dong Liu, and Jingdong Wang,
\newblock ``Deep high-resolution representation learning for human pose
  estimation,''
\newblock in {\em Proceedings of the IEEE/CVF Conference on Computer Vision and
  Pattern Recognition}, 2019, pp. 5693--5703.

\bibitem{kong2020sia}
Deying Kong, Haoyu Ma, and Xiaohui Xie,
\newblock ``Sia-gcn: A spatial information aware graph neural network with 2d
  convolutions for hand pose estimation,''
\newblock {\em arXiv preprint arXiv:2009.12473}, 2020.

\bibitem{chen2020nonparametric}
Yifei Chen, Haoyu Ma, Deying Kong, Xiangyi Yan, Jianbao Wu, Wei Fan, and
  Xiaohui Xie,
\newblock ``Nonparametric structure regularization machine for 2d hand pose
  estimation,''
\newblock in {\em Proceedings of the IEEE/CVF Winter Conference on Applications
  of Computer Vision}, 2020, pp. 381--390.

\bibitem{wang2019geometric}
Zhe Wang, Liyan Chen, Shaurya Rathore, Daeyun Shin, and Charless Fowlkes,
\newblock ``Geometric pose affordance: 3d human pose with scene constraints,''
\newblock {\em arXiv preprint arXiv:1905.07718}, 2019.

\bibitem{Ren2015FasterRT}
Shaoqing Ren, Kaiming He, Ross~B. Girshick, and Jian Sun,
\newblock ``Faster r-cnn: Towards real-time object detection with region
  proposal networks,''
\newblock {\em IEEE Trans. on Pattern Analysis and Machine Intelligence}, vol.
  39, pp. 1137--1149, 2015.

\bibitem{Xiao2018VideoOD}
Fanyi Xiao and Yong~Jae Lee,
\newblock ``Video object detection with an aligned spatial-temporal memory,''
\newblock {\em in Proc. of the European Conference on Computer Vision}, 2018.

\bibitem{wang2019learning}
Xiaolong Wang, Allan Jabri, and Alexei~A Efros,
\newblock ``Learning correspondence from the cycle-consistency of time,''
\newblock in {\em Proceedings of the IEEE Conference on Computer Vision and
  Pattern Recognition}, 2019, pp. 2566--2576.

\bibitem{Han2016SeqNMSFV}
Wei Han, Pooya Khorrami, Tom~Le Paine, Prajit Ramachandran, Mohammad
  Babaeizadeh, Humphrey Shi, Jianan Li, Shuicheng Yan, and Thomas~S. Huang,
\newblock ``Seq-nms for video object detection,''
\newblock {\em ArXiv}, vol. abs/1602.08465, 2016.

\bibitem{Zhu2017DeepFF}
Xizhou Zhu, Yuwen Xiong, Jifeng Dai, Lu~Yuan, and Yichen Wei,
\newblock ``Deep feature flow for video recognition,''
\newblock {\em in Proc. of IEEE Conference Computer Vision and Pattern
  Recognition}, pp. 4141--4150, 2017.

\bibitem{Hetang2017ImpressionNF}
Congrui Hetang, Hongwei Qin, Shaohui Liu, and Junjie Yan,
\newblock ``Impression network for video object detection,''
\newblock {\em ArXiv}, vol. abs/1712.05896, 2017.

\bibitem{Donahue2015LongtermRC}
Jeff Donahue, Lisa~Anne Hendricks, Marcus Rohrbach, Subhashini Venugopalan,
  Sergio Guadarrama, Kate Saenko, and Trevor Darrell,
\newblock ``Long-term recurrent convolutional networks for visual recognition
  and description,''
\newblock {\em in Proc. of IEEE Conference Computer Vision and Pattern
  Recognition}, pp. 2625--2634, 2015.

\bibitem{Lu2017OnlineVO}
Yongyi Lu, Cewu Lu, and Chi-Keung Tang,
\newblock ``Online video object detection using association lstm,''
\newblock {\em in Proc. of IEEE Conference International Conference on Computer
  Vision}, pp. 2363--2371, 2017.

\bibitem{Ning2017SpatiallySR}
Guanghan Ning, Zhi Zhang, Chen Huang, Xiaobo Ren, Haohong Wang, Canhui Cai, and
  Zhihai He,
\newblock ``Spatially supervised recurrent convolutional neural networks for
  visual object tracking,''
\newblock {\em in Proc. of IEEE International Symposium on Circuits and
  Systems}, pp. 1--4, 2017.

\bibitem{Liu2018MobileVO}
Mason Liu and Menglong Zhu,
\newblock ``Mobile video object detection with temporally-aware feature maps,''
\newblock {\em in Proc. of IEEE Conference Computer Vision and Pattern
  Recognition}, pp. 5686--5695, 2018.

\bibitem{siam2017convolutional}
Mennatullah Siam, Sepehr Valipour, Martin Jagersand, and Nilanjan Ray,
\newblock ``Convolutional gated recurrent networks for video segmentation,''
\newblock in {\em 2017 IEEE International Conference on Image Processing
  (ICIP)}. IEEE, 2017, pp. 3090--3094.

\bibitem{chung2014empirical}
Junyoung Chung, Caglar Gulcehre, KyungHyun Cho, and Yoshua Bengio,
\newblock ``Empirical evaluation of gated recurrent neural networks on sequence
  modeling,''
\newblock {\em arXiv preprint arXiv:1412.3555}, 2014.

\bibitem{hochreiter1997long}
Sepp Hochreiter and J{\"u}rgen Schmidhuber,
\newblock ``Long short-term memory,''
\newblock {\em Neural computation}, vol. 9, no. 8, pp. 1735--1780, 1997.

\bibitem{Howard2017MobileNetsEC}
Andrew~G. Howard, Menglong Zhu, Bo~Chen, Dmitry Kalenichenko, Weijun Wang,
  Tobias Weyand, Marco Andreetto, and Hartwig Adam,
\newblock ``Mobilenets: Efficient convolutional neural networks for mobile
  vision applications,''
\newblock {\em ArXiv}, vol. abs/1704.04861, 2017.

\bibitem{Liu2016SSDSS}
Wei Liu, Dragomir Anguelov, Dumitru Erhan, Christian Szegedy, Scott~E. Reed,
  Cheng-Yang Fu, and Alexander~C. Berg,
\newblock ``Ssd: Single shot multibox detector,''
\newblock {\em in Proc. of the European Conference on Computer Vision}, 2016.

\bibitem{He2016DeepRL}
Kaiming He, Xiangyu Zhang, Shaoqing Ren, and Jian Sun,
\newblock ``Deep residual learning for image recognition,''
\newblock {\em in Proc. of IEEE Conference Computer Vision and Pattern
  Recognition}, pp. 770--778, 2016.

\bibitem{Deng2009ImageNetAL}
Jia Deng, Wenjun Dong, Richard Socher, Li-Jia Li, Kai Li, and Li~Fei-Fei,
\newblock ``Imagenet: A large-scale hierarchical image database,''
\newblock {\em in Proc. of IEEE Conference Computer Vision and Pattern
  Recognition}, 2009.

\end{thebibliography}

\end{document}